\documentclass[conference]{IEEEtran}
\IEEEoverridecommandlockouts
\usepackage{cite}
\usepackage{amsmath,amssymb,amsfonts}
\usepackage{algorithmic}
\usepackage{graphicx}
\usepackage{textcomp}
\usepackage{multirow}
\usepackage[table,xcdraw]{xcolor}
\usepackage{csquotes}
\usepackage{float}

\begin{document}

\title{QEMesh: Employing A Quadric Error Metrics-Based Representation for Mesh Generation}

\author{Jiaqi Li$^{1}$, Ruowei Wang$^{2}$, Yu Liu$^{1}$, Qijun Zhao$^{1,2,*}$\thanks{* Corresponding author.} \\
$^{1}$National Key Laboratory of Fundamental Science on Synthetic Vision, Sichuan University, Chengdu, China\\
$^{2}$College of Computer Science Sichuan University, Chengdu, China \\
{slrrr8848@gmail.com;
 ruoweiwang@stu.scu.edu.cn; liuyuvincent@stu.scu.edu.cn; qjzhao@scu.edu.cn}
}

\maketitle

\begin{abstract}
Mesh generation plays a crucial role in 3D content creation, as mesh is widely used in various industrial applications. Recent works have achieved impressive results but still face several issues, such as unrealistic patterns or pits on surfaces, thin parts missing, and incomplete structures. Most of these problems stem from the choice of shape representation or the capabilities of the generative network. To alleviate these, we extend PoNQ, a Quadric Error Metrics (QEM)-based representation, and propose a novel model, QEMesh, for high-quality mesh generation. PoNQ divides the shape surface into tiny patches, each represented by a point with its normal and QEM matrix, which preserves fine local geometry information. In our QEMesh, we regard these elements as generable parameters and design a unique latent diffusion model containing a novel multi-decoder VAE for PoNQ parameters generation. Given the latent code generated by the diffusion model, three parameter decoders produce several PoNQ parameters within each voxel cell, and an occupancy decoder predicts which voxel cells containing parameters to form the final shape. Extensive evaluations demonstrate that our method generates results with watertight surfaces and is comparable to state-of-the-art methods in several main metrics.

\end{abstract}


\section{Introduction}
\label{sec:intro}
Mesh generation is a principal task in 3D vision, as mesh is the most widely used representation of 3D shapes due to its versatility in rendering. Meshes with continuous, watertight surfaces ensure smooth transitions and prevent geometric artifacts. These properties are essential for 3D rendering, texturing, and simulation. However, the irregular data structure and varying topologies of meshes pose challenges for neural networks in processing them directly.

Traditional methods utilize point clouds \cite{vahdat2022lion,getmesh, slide, luodiffusion}, 3D voxels \cite{pvdiffusion, occupancyNet,li2023voxformer}, and implicit functions\cite{autosdf, deepsdf, lasdiffusion, sdf_diffusion1, sdfstylegan} as intermediate representations, which are easy to obtain and more compatible with neural networks. 
They require additional reconstruction methods to extract meshes, such as MC \cite{chernyaev1995marchingcubes33} and SAP \cite{peng2021shape}, which leads to a loss of quality, such as unrealistic surfaces.
Another group of works \cite{meshgpt, polydiff, nash2020polygen, chen2024meshxl} break down complex meshes into more manageable components before handing them over to neural networks. They treat vertices or faces as sequences with a transformer network \cite{transformer} and generate them step by step as large language models do.
\begin{figure}[h]
    \centering
    \includegraphics[width=1\linewidth]{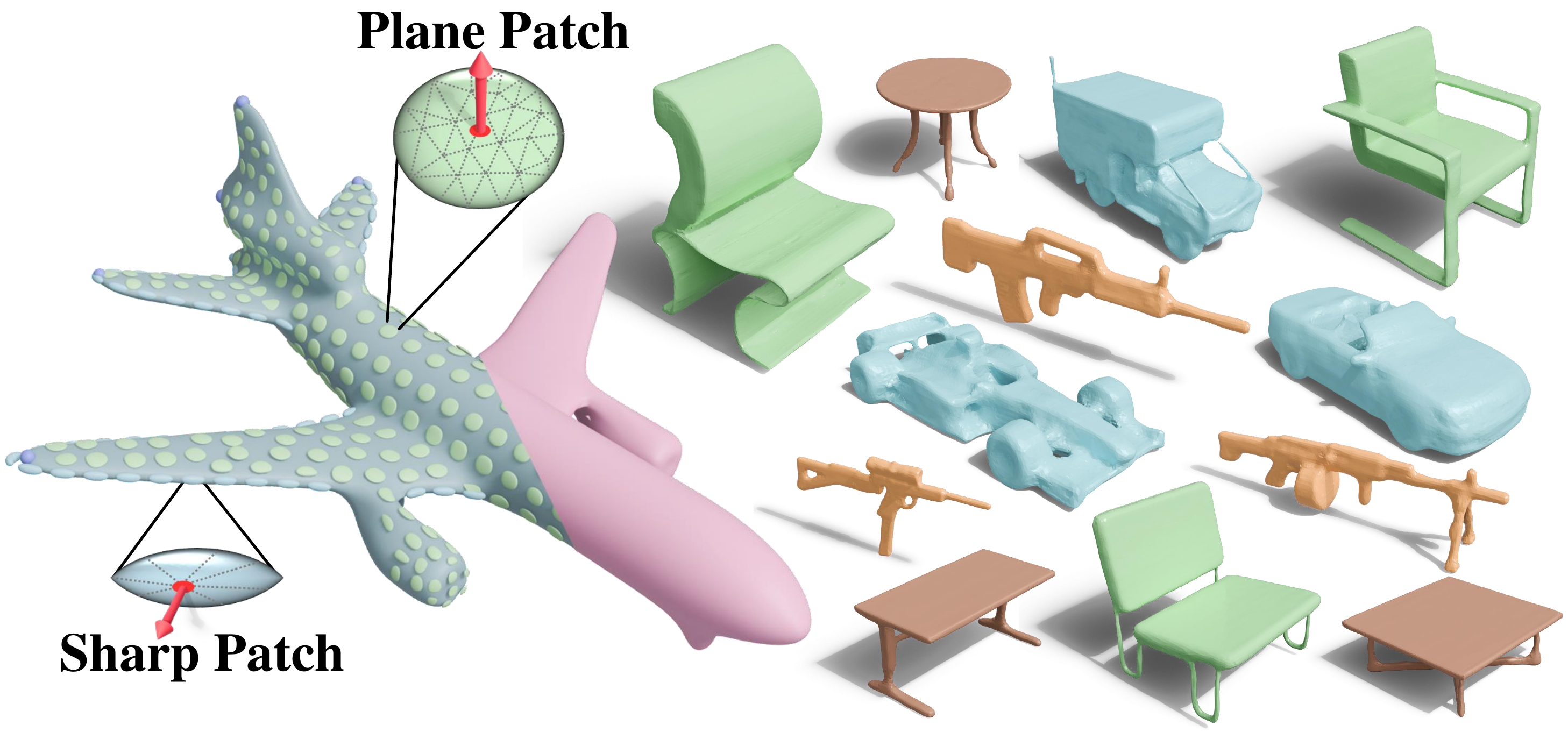}

    \caption{
    (Left) Visualization of PoNQ representation on the airplane, in which the Quadric error metric matrix $Q_{i}$ encodes faces of a small surface patch. The patches in different shapes indicate different local geometric information. (Right) We also place some samples from our method at $64^3$ resolution.
    }
    \label{fig:F1}
\end{figure}
However, as sequence length increases, these methods demand more resources and struggle to generate complex structures, resulting in meshes with fewer faces as mentioned in \cite{GenUDC}.
MeshDiffusion\cite{meshdiffusion} combines a deformable tetrahedral grid with Signed Distance Functions (SDF) to represent meshes. It relies on 2D supervision to predict vertex positions and SDF values. However, the results often exhibit noticeable distortions or wrinkles due to inaccurate predictions.
Inspired by the idea of Dual Contouring\cite{dc}, GenUDC\cite{GenUDC} proposes the UDC representation, which splits a mesh into faces part and vertices part in a regular grid. This decomposition makes it convenient for neural networks to process. However, GenUDC cannot guarantee the results with watertight surfaces and suffers from the non-manifold problem. All these limitations of previous representation or model designs highlight the demand for methods capable of generating meshes that capture intricate details and guarantee watertight surfaces suitable for real-world applications. 

In this paper, we propose a novel surface patch-parametrized model, QEMesh, which explores the potential of PoNQ representation \cite{ponq} in the mesh generation task. We refer to the PoNQ representation as PoNQ with a little abuse of the abbreviation.
In PoNQ, the shape surface is divided into tiny patches as shown in Fig.~\ref{fig:F1}, each represented by a point with its normal and Quadric Error Metrics (QEM) matrix, all of which we regard as generable parameters.
QEMesh consists of a multi-decoder VAE that decodes PoNQ parameters and a diffusion model that learns the probability distribution of the latent PoNQ shape representation. The diffusion model first recovers a random noise to a latent PoNQ code. Then, three parameter decoders decode it to PoNQ parameters in a regular grid and an occupancy decoder filters the parameters that contribute to the shape. Specifically, the occupancy decoder takes the latent code and grid cell coordinates as input to predict whether the parameters within the grid cell are a part of the final shape or just noise. In this way, we achieve more efficient and accurate mask prediction. The holistic framework of QEMesh is pictured in Fig.~\ref{fig:pipeline}.


Our contributions can be summarized below:
\begin{itemize}
\item 
We propose a novel framework dubbed QEMesh for high-quality mesh generation which for the first time utilizes PoNQ representation.
\item We develop a multi-decoder VAE for PoNQ generation, which regards a surface patch as a generable parameter.
\item Our occupancy decoder leverages a point-wise strategy for accurate mask prediction, which is compatible with the QEMesh framework.
\item Extensive experiments validate the superiority of QEMesh over existing models in mesh generation, and further ablation studies also show that our design in the VAE is well-suited.
\end{itemize}
\section{Related Works}
\subsection{Shape Representation in 3D Generation}
Early explorations of 3D generation focused on explicit representations such as point clouds \cite{vahdat2022lion,luodiffusion,slide, getmesh} and 3D voxels \cite{pvdiffusion,SALAD} due to their straightforward shape representation. However, methods utilizing point clouds require a fixed number of points as inputs and outputs, which has limitations in capturing complex structures, and 3D voxels are computationally expensive, which hinders their development. Recent progress in neural implicit representation \cite{im_gan, occupancyNet,deepsdf}, such as occupancy value and SDF, enable efficient 3D shape learning. However, these methods mainly generate values at fixed positions and need additional post-processing steps, which limits their performances.

A group of concurrent works \cite{chen2024textgs,yi2024gaussiandreamer,xu2024grm} utilizes NeRF or 3D Gaussian Splatting (3DGS) as 3D representations. Following an optimization-based process via generative models, the resulting NeRF or 3DGS contains shape and texture information. However, the post-processed results may be plagued by artifacts such as detail loss and blurriness.

Furthermore, some works seek an effective mesh representation for mesh generation. These methods \cite{nash2020polygen}, \cite{meshgpt}, \cite{polydiff}, \cite{chen2024meshxl} construct a mesh codebook to learn a set of shapes. They recall meshes in the codebook to assemble a shape.
However, they cannot produce complex structures due to memory limits. Some other works try to integrate meshes with grids. MeshDiffusion\cite{meshdiffusion} utilizes a deformable tetrahedral grid and SDF to generate meshes. It computes mesh vertices via linear interpolation from the tetrahedral grid and SDF values, then connects vertices to form faces within the same tetrahedrons. However, as noted in \cite{GenUDC}, its reliance on 2D supervision results in inaccurate reconstructions. GenUDC \cite{GenUDC} introduces the UDC representation, which discretizes a mesh in a regular grid, making it easier to handle in neural networks. However, it inherits non-manifold issues of dual contouring (DC) \cite{dc}, and is prone to generating holes on the surface. Our work uses PoNQ \cite{ponq} to facilitate the mesh generation task. PoNQ encodes a mesh through a set of points, normals, and QEM matrices \cite{qem1997}. We treat them as equivalent to parameters and generate them within a regular voxel grid.
\subsection{Diffusion Models}
Unlike classic generative models such as generative adversarial networks (GAN) \cite{gan} and variational autoencoders (VAE) \cite{vae}, which generate samples in a single step, diffusion models employ an iterative generation process, progressively refining the output at each step. It learns data distribution by utilizing a denoising strategy, enabling the model to map a sample of simple distribution (e.g., Gaussian distribution) to a complex data distribution. The recent success of diffusion models in 2D image generation has sparked significant interest in adapting these techniques for 3D content \cite{GenUDC,sdf_diffusion1,luo2021diffusion,slide,meshdiffusion,JIJIN1, JIJIN2, JIJIN3}. Following the pioneering work, DDPM \cite{ho2020denoisingddpm}, researchers began to explore its application to existing 3D shape representations. However, most early methods operate directly in 3D space, resulting in huge storage demands and slow performance during training and inference. Subsequent work, the Latent Diffusion Model (LDM) \cite{rombach2022highldm}, proposed training diffusion models in a lower-dimensional latent space constructed by VAE, which speeds up training and reduces memory demands while not sacrificing generation quality. In our work, we devise a novel LDM containing a multi-decoder VAE for PoNQ generation.

\begin{figure*}[ht]
    \centering
    \includegraphics[width=1\linewidth]{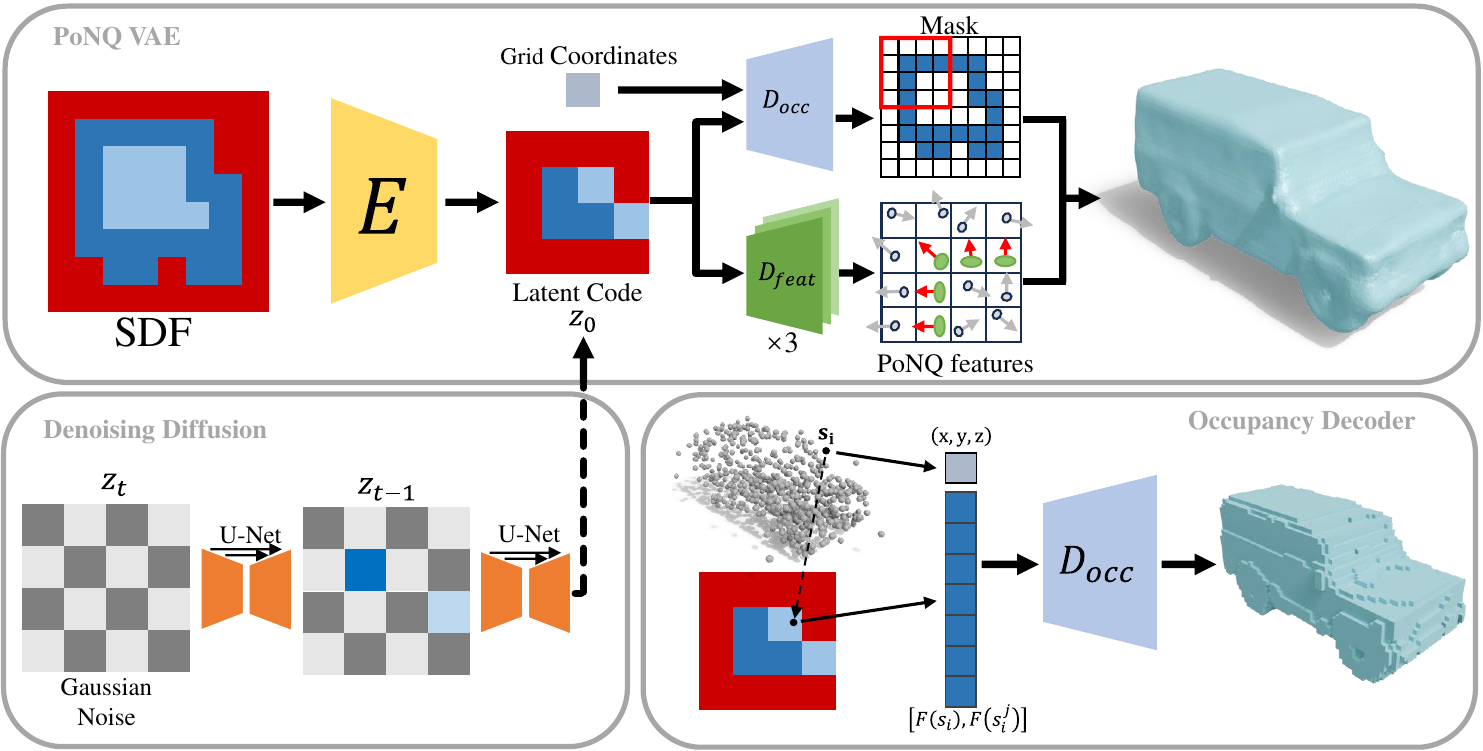}

    \caption{
    An overview of QEMesh: We show the case where \textbf{each grid holds only $\textbf{K=1}$ point, normal, and $Q$ matrix}. The denoising diffusion step removes the noise to obtain a latent code. Then the parameter decoders convert it to PoNQ parameters (ellipses represent $\hat{v_i}$ and $Q_i$ and arrows represent the normal), and the occupancy decoder produces a mask to select parameters relevant to a shape (green ellipses). To train the occupancy decoder, $D_{occ}$ takes a point $s_i$ and latent features as the inputs to predict whether a cell is occupied by the shape. The blue squares are the features of $s_i$ and its six surrounding points.
    }
    \label{fig:pipeline}
\end{figure*}

\section{Method}

\subsection{Overview}
The main challenges of the mesh generation task lie in modeling shapes with a good 3d representation and leveraging neural networks to learn the distribution of 3d representation. To address them, we extend the PoNQ representation to mesh generation and propose the QEMesh framework. 
To represent a shape in PoNQ, a small surface patch is parametrized by a single point with its normal and Quadric Error Metrics matrix $Q$, which approximates the faces within the patch as shown in Fig.~\ref{fig:F1}. In QEMesh, we develop a multi-decoder VAE that generates PoNQ parameters in a voxel grid with the size of $(N, N, N)$ as shown in Fig.~\ref{fig:pipeline}. It consists of two parts: three parameter decoders that generate PoNQ parameters inside the grid and an occupancy decoder following a point-wise strategy to select parameters relevant to the shape. 




\subsection{PoNQ Representation}
\label{PoNQ}
\textbf{Quadric Error Metrics.} Given a tangent plane $p=ax+by+cz$ ($a^2+b^2+c^2=1$) with its normal $n_p=[a,b,c,d]^T$, the squared distance from a given position $v=[x,y,z,1]^T$ to $p$ can be defined as 
\begin{align}
    d_{p}(v,n_p) & = (v^Tn_p)^2 \\ &=v^Tn_pn_p^Tv \\ &=v^TK_{p}v,
\end{align} where $K_{p}=n_pn_p^T$. Then the squared distance from $v$ to a plane set $Planes_i= \{p_1,p_2,...,p_n\}$ can be formed as 
\begin{align}
    \operatorname{QEM}(v) & = \sum_{p \in Planes_i} d_{p}(v,n_i) \\ & = v^T(\sum_{p \in Planes_i}K_{p})v \\ & = v^TQ_i v,
\end{align}
where $Q_i=\sum_{p \in Planes_i}K_{p}$. In the scenario of mesh decimation, $Planes_i$ represents \textbf{the triangle faces of a tiny patch}. By minimizing $\operatorname{QEM}(v)$ we obtain the optimal position
\begin{equation}
    \hat{v_i} =\underset{v}{\operatorname{argmin}v^TQv},
    \label{v_opt}
\end{equation}
which replaces some of the triangle faces' vertices after an iterative vertex merging to reduce the number of faces.

\textbf{Shape Representation.} According to \cite{ponq}, the matrix $Q$ can represent the union of adjacent tangent planes $Planes_i$, i.e., mesh faces. Specifically, a $\hat{v_i}$ with its $Q_i$ can represent an approximation of a small local region. Thus, a shape can be represented with the point set \(\mathbf{P}=\{\hat{v_i}\in \mathbb{R}^3\}\) and the matrix set \(\mathbf{Q}=\{Q_i\in \mathbb{R}^{4\times4}\}\). The normal set \(\mathbf{N}=\{n_i\in \mathbb{R}^3\}\) is still needed and derived by directly averaging the normals in the small region. When using computational geometry methods\cite{ponq,amenta1998newvoronoi} to extract meshes, all three sets: \(\mathbf{P}\), \(\mathbf{N}\), \(\mathbf{Q}\) are used.

\subsection{PoNQ VAE}
\label{PVAE}
We devise a multi-decoder VAE for PoNQ parameters generation. It consists of an encoder $E$, an occupancy decoder $D_{occ}$ and three parameter decoders 
$D_{feat}=\{D_{\mathbf{P}},D_{\mathbf{N}},D_{\mathbf{Q}}\}$. Given an input $X$ represented via SDF, $E$ encodes $X$ to a latent representation $z=E(X)$. Taking $z$ as input, $D_{feat}$ converts it to PoNQ parameters \(\mathbf{P}\), \(\mathbf{N}\), \(\mathbf{Q}\) while $D_{occ}$ converts it to a boolean mask with extra input of grid cell coordinates. The parameters within cells marked as $1$ are utilized for the final shape representation. In the following part, we provide a detailed illustration of our decoders.
\begin{figure*}[htbp]
    \centering
    \includegraphics[width=0.9\linewidth]{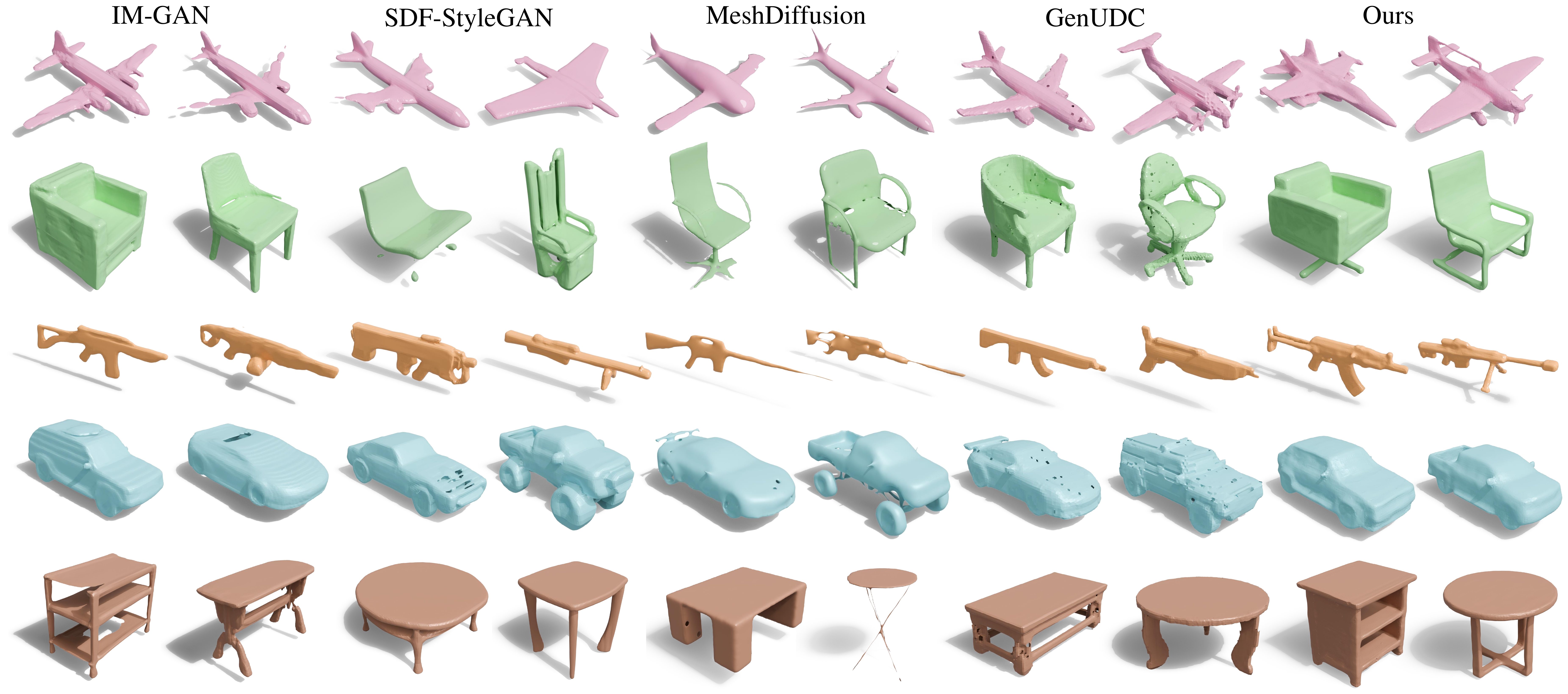}

    \caption{
    Qualitative evaluation of shape generation at $64^3$ resolution. 
    }
    \label{fig:visual}
\end{figure*}



\textbf{Parameter Decoders} To train our parameter decoders, we follow PoNQ \cite{ponq} for data preparation: 1. Densly sample points $s_j \in S$ and normals $n(s_j)$ from mesh surfaces $S$ to represent mesh faces; 2. Discretize $S$ in a regular grid to split the shape into patches $P(x_i)$, where $x_i$ denotes the center of the grid. $D_{feat}$ will generate $\textbf{K}$ points with their normals and matrices $Q$ per grid cell which is supervised by $L_{feat}=\alpha L_{\hat{v}}+ \beta L_{n} + \gamma L_{Q}$, where
\begin{align}
& L_{\hat{v}}= \frac{1}{n} \sum_{i=1}^n \sum_{s_j \in S \cap P\left(x_i\right)}\left(\hat{v}_in\left(s_j\right)^t\right)^2,\ \\ 
& L_{n}= \frac{1}{n} \sum_{i=1}^n \sum_{s_j \in S \cap P\left(x_i\right)}\left\|n_i-n\left(s_j\right)\right\|^2 , \\
& L_{Q}= \frac{1}{n} \sum_{i=1}^n \sum_{s_j \in S \cap P\left(x_i\right)}\left\|Q_i-n\left(s_j\right)^t n\left(s_j\right)\right\|^2.
\end{align}
At inference time, PoNQ parameters are combined with the mask generated by $D_{occ}$ to form the final shape. Notice that the value of $\textbf{K}$ depends on the complexity of the structure in different categories.

\textbf{Occupancy Decoder} Incorrect parameter selection will lead to incomplete structures, surface artifacts, or noise in the space, as shown in Fig.~\ref{fig:ablation}.
Inspired by \cite{im_gan}, $D_{occ}$ takes the coordinates of a grid cell center and the latent code $z = E(X)\in \mathbb{R}^{c\times n\times n\times n}$, which is seen as a latent code grid with $N$ resolution to determine whether a cell contains parameters that form a part of the surface. This can be regarded as generating a coarse shape made of voxels. To create training point samples, we first sample a number of points $s_i\in S$ on the surface and add random displacements to $s_i$ to make them spread around the space. Then we compute the ground truth occupancy $O_{gt}(s_i)\in(0,1)$, which equals 1 for $s_i$ inside the voxel containing the surface and 0 otherwise. In the training stage, $D_{occ}$ first extracts the latent feature $F(s_i) \in z$ from the grid at position $s_i$ using trilinear interpolation. To capture the local context, we gather latent features $F(s_i^j)\in z$ from six neighboring points $s_i^j$. Then we feed the concatenated $\mathbf{Z}=[F(s_i),F(s_i^j),s_i]$ to $D_{occ}$ to get the occupancy value of the voxel containing $p$. We show the framework of $D_{occ}$ in Fig.~\ref{fig:pipeline}.
By evaluating the occupancy of the voxels, we bypass the need to individually assess each point in the space.

\subsection{Diffusion Model}
\label{DM}
We use the latent code $z=E(X)$ to train the diffusion model to learn a distribution $p(z|X)$. We first normalize $z_0\in p(z|X)$ to [-1.0,1.0] by max-min normalization. In the forward process, a controlled Gaussian noise $\epsilon \sim N(0,1)$ is gradually added to $z_0$ to form $z_t=\sqrt{\bar{\alpha}_t} z_0+\sqrt{1-\bar{\alpha}_t} \boldsymbol{\epsilon}$, where $t=1,2,...,T$ and $\bar{\alpha}_t $ is a decreasing function from 1 to 0. In the reverse process, the prediction from $z_t$ to $z_0$ can be modeled using a 3D U-Net $\boldsymbol{\epsilon}(z_t,t)$ following:
\begin{align}
     L_{d m}=\mathbb{E}_{\boldsymbol{z}, t, \epsilon \sim N(0,1)}\left\|\boldsymbol{\epsilon}-\boldsymbol{\epsilon}_{\boldsymbol{\theta}}\left(z_t, t\right)\right\|_1.
\end{align}
In the inference stage, the diffusion model denoises a random sampled Gaussian noise $z_t=\epsilon \sim N(0,1)$ to $z_0$. $D_{para}$ and $D_{occ}$ take the denormalized $z_0$ as input and decode it to the corresponding results. 

\section{Experiments}
\subsection{Data Preparation}
Following GenUDC's setup \cite{GenUDC}, we use the ShapeNet Core (v1) dataset \cite{shapenet2015} to train and evaluate our model, focusing on five categories: airplane, car, chair, rifle, and table. The dataset is split into $70\%$ for training, $20\%$ for testing, and $10\%$ for validation, with the validation set unused.

\subsection{Unconditional Shape Generation}

\begin{table*}[t]
\caption{
Quantitative evaluation of shape generation at $64^3$ resolution. 
}
\centering
\resizebox{0.9\linewidth}{!}{
\begin{tabular}{clccc|ccc|ccc|c|c|c}
\hline
 &  & \multicolumn{3}{c|}{MMD ($\downarrow$)} & \multicolumn{3}{c|}{COV ($\%, \uparrow$)} & \multicolumn{3}{c|}{1-NNA ($\%, \downarrow$)} &  &  & \multicolumn{1}{l}{} \\ \cline{3-11}
\multirow{-2}{*}{} & \multirow{-2}{*}{Method} & CD $\times 10^3$ & EMD $\times 10$ & LFD & CD & EMD & LFD & CD & EMD & LFD & \multirow{-2}{*}{JSD $\times 10^3 (\downarrow)$} & \multirow{-2}{*}{\begin{tabular}[c]{@{}c@{}}Watertight\\ ($\%, \uparrow$)\end{tabular}} & \multicolumn{1}{l}{\multirow{-2}{*}{\begin{tabular}[c]{@{}l@{}}Self-int.\\ ($\%, \downarrow$)\end{tabular}}} \\ \hline
\multicolumn{1}{c|}{} & IM-GAN & 3.736 & 1.110 & 4939 & 44.25 & 37.08 & {\color[HTML]{333333} \textbf{45.86}} & 79.48 & 82.94 & 79.11 & 21.151 & \textbf{100} & \cellcolor[HTML]{FFFFFF}\textbf{0} \\
\multicolumn{1}{c|}{} & SDF-StyleGAN & 4.558 & 1.180 & 5326 & 40.67 & 32.63 & 38.2 & 85.48 & 87.08 & 84.73 & 26.304 & \textbf{100} & \cellcolor[HTML]{FFFFFF}\textbf{0} \\
\multicolumn{1}{c|}{} & MeshDiffusion & {\color[HTML]{333333} \textbf{3.612}} & 1.042 & 4538 & 47.34 & 42.15 & 45.36 & 66.44 & 76.26 & {\color[HTML]{333333} \textbf{67.24}} & 11.366 & 26.67 & 73.33 \\
\multicolumn{1}{c|}{} & GenUDC & 3.960 & {\color[HTML]{333333} 0.902} & {\color[HTML]{333333} 3167} & {\color[HTML]{333333} 48.33} & {\color[HTML]{333333} 50.06} & 44.13 & {\color[HTML]{333333} 60.75} & {\color[HTML]{333333} 56.74} & 69.16 & {\color[HTML]{333333} \textbf{7.020}} & 0.49 & 28.33 \\
\multicolumn{1}{c|}{\multirow{-5}{*}{Airplane}} & \cellcolor[HTML]{EFEFEF}\textbf{Ours} & \cellcolor[HTML]{EFEFEF}3.751 & \cellcolor[HTML]{EFEFEF}{\color[HTML]{000000} \textbf{0.827}} & \cellcolor[HTML]{EFEFEF}{\color[HTML]{000000} \textbf{3007}} & \cellcolor[HTML]{EFEFEF}{\color[HTML]{000000} \textbf{49.82}} & \cellcolor[HTML]{EFEFEF}{\color[HTML]{000000} \textbf{50.95}} & \cellcolor[HTML]{EFEFEF}44.91 & \cellcolor[HTML]{EFEFEF}{\color[HTML]{000000} \textbf{60.01}} & \cellcolor[HTML]{EFEFEF}\textbf{55.15} & \cellcolor[HTML]{EFEFEF}69.26 & \cellcolor[HTML]{EFEFEF}7.566 & \cellcolor[HTML]{EFEFEF}\textbf{100} & \cellcolor[HTML]{EFEFEF}\textbf{0} \\ \hline
\multicolumn{1}{c|}{} & IM-GAN & 5.209 & 1.197 & 2645 & 28.26 & 24.92 & 30.73 & 95.69 & 94.79 & 89.3 & 42.586 & \textbf{100} & \cellcolor[HTML]{FFFFFF}\textbf{0} \\
\multicolumn{1}{c|}{} & SDF-StyleGAN & 5.064 & 1.152 & 2623 & 29.93 & 32.06 & 41.93 & 88.34 & 88.31 & 84.13 & 15.960 & \textbf{100} & \cellcolor[HTML]{FFFFFF}\textbf{0} \\
\multicolumn{1}{c|}{} & MeshDiffusion & 4.972 & 1.196 & 2477 & 34.07 & 25.85 & 37.53 & 81.43 & 87.84 & 70.83 & 12.384 & 0 & 100 \\
\multicolumn{1}{c|}{} & GenUDC & 3.753 & {\color[HTML]{333333} 0.854} & {\color[HTML]{333333} 1191} & {\color[HTML]{333333} 45.67} & {\color[HTML]{333333} \textbf{46.53}} & {\color[HTML]{333333} 45.73} & {\color[HTML]{333333} 60.80} & {\color[HTML]{333333} 58.33} & {\color[HTML]{333333} \textbf{62.23}} & {\color[HTML]{333333} \textbf{2.839}} & 0 & 66.33 \\
\multicolumn{1}{c|}{\multirow{-5}{*}{Car}} & \cellcolor[HTML]{EFEFEF}\textbf{Ours} & \cellcolor[HTML]{F2F2F2}{\color[HTML]{000000} \textbf{3.506}} & \cellcolor[HTML]{EFEFEF}{\color[HTML]{000000} \textbf{0.813}} & \cellcolor[HTML]{EFEFEF}{\color[HTML]{000000} \textbf{1068}} & \cellcolor[HTML]{EFEFEF}\textbf{46.93} & \cellcolor[HTML]{EFEFEF}45.95 & \cellcolor[HTML]{EFEFEF}\textbf{46.60} & \cellcolor[HTML]{F2F2F2}{\color[HTML]{000000} \textbf{58.40}} & \cellcolor[HTML]{EFEFEF}\textbf{56.38} & \cellcolor[HTML]{EFEFEF}62.83 & \cellcolor[HTML]{EFEFEF}3.102 & \cellcolor[HTML]{EFEFEF}\textbf{100} & \cellcolor[HTML]{EFEFEF}\textbf{0} \\ \hline
\multicolumn{1}{c|}{} & IM-GAN & {\color[HTML]{333333} 11.378} & 1.567 & 3400 & {\color[HTML]{333333} 51.04} & 49.20 & 51.04 & 65.96 & 63.17 & 62.49 & 4.865 & \textbf{100} & \cellcolor[HTML]{FFFFFF}\textbf{0} \\
\multicolumn{1}{c|}{} & SDF-StyleGAN & 13.896 & 1.615 & 3423 & 42.21 & 41.80 & 42.98 & 68.35 & 68.21 & 66.19 & 4.603 & \textbf{100} & \cellcolor[HTML]{FFFFFF}\textbf{0} \\
\multicolumn{1}{c|}{} & MeshDiffusion & 11.405 & 1.548 & 3427 & 49.56 & 50.33 & {\color[HTML]{333333} \textbf{51.92}} & 59.35 & 59.47 & {\color[HTML]{333333} \textbf{58.97}} & 4.310 & 36.33 & 57.13 \\
\multicolumn{1}{c|}{} & GenUDC & 11.998 & 1.564 & 2683 & 46.36 & {\color[HTML]{333333} 50.41} & 47.12 & 61.46 & 59.43 & 60.75 & 3.822 & 2.17 & 24.85 \\
\multicolumn{1}{c|}{\multirow{-5}{*}{Table}} & \cellcolor[HTML]{EFEFEF}\textbf{Ours} & \cellcolor[HTML]{EFEFEF}{\color[HTML]{000000} \textbf{11.078}} & \cellcolor[HTML]{EFEFEF}{\color[HTML]{000000} \textbf{1.461}} & \cellcolor[HTML]{EFEFEF}{\color[HTML]{000000} \textbf{2602}} & \cellcolor[HTML]{EFEFEF}{\color[HTML]{000000} \textbf{51.94}} & \cellcolor[HTML]{EFEFEF}{\color[HTML]{000000} \textbf{50.88}} & \cellcolor[HTML]{EFEFEF}50.76 & \cellcolor[HTML]{EFEFEF}{\color[HTML]{000000} \textbf{57.62}} & \cellcolor[HTML]{EFEFEF}{\color[HTML]{000000} \textbf{57.23}} & \cellcolor[HTML]{EFEFEF}59.22 & \cellcolor[HTML]{EFEFEF}{\color[HTML]{000000} \textbf{3.103}} & \cellcolor[HTML]{EFEFEF}\textbf{100} & \cellcolor[HTML]{EFEFEF}\textbf{0} \\ \hline
\multicolumn{1}{c|}{} & IM-GAN & 3.550 & 1.058 & 6240 & 46.53 & 37.89 & 42.32 & 70.00 & 72.74 & 69.26 & 25.704 & \textbf{100} & \cellcolor[HTML]{FFFFFF}\textbf{0} \\
\multicolumn{1}{c|}{} & SDF-StyleGAN & 4.100 & 1.069 & 6475 & 46.53 & 40.21 & 41.47 & 73.68 & 73.16 & 76.84 & 33.624 & \textbf{100} & \cellcolor[HTML]{FFFFFF}\textbf{0} \\
\multicolumn{1}{c|}{} & MeshDiffusion & {\color[HTML]{333333} \textbf{3.124}} & 1.018 & 5951 & {\color[HTML]{333333} \textbf{52.63}} & 42.11 & 48.84 & 57.68 & 67.79 & \textbf{55.58} & 19.353 & 73.68 & 29.47 \\
\multicolumn{1}{c|}{} & GenUDC & 3.530 & 0.849 & 3493 & 48.42 & {\color[HTML]{333333} 51.58} & \textbf{50.53} & {\color[HTML]{333333} 56.63} & {\color[HTML]{333333} 55.05} & \textbf{55.58} & {\color[HTML]{333333} \textbf{10.951}} & 1.89 & 9.26 \\
\multicolumn{1}{c|}{\multirow{-5}{*}{Rifle}} & \cellcolor[HTML]{EFEFEF}\textbf{Ours} & \cellcolor[HTML]{EFEFEF}3.600 & \cellcolor[HTML]{EFEFEF}{\color[HTML]{000000} \textbf{0.792}} & \cellcolor[HTML]{EFEFEF}{\color[HTML]{000000} \textbf{3432}} & \cellcolor[HTML]{EFEFEF}49.21 & \cellcolor[HTML]{EFEFEF}\textbf{52.32} & \cellcolor[HTML]{EFEFEF}48.63 & \cellcolor[HTML]{EFEFEF}{\color[HTML]{000000} \textbf{55.36}} & \cellcolor[HTML]{EFEFEF}{\color[HTML]{000000} \textbf{54.84}} & \cellcolor[HTML]{EFEFEF}56.52 & \cellcolor[HTML]{EFEFEF}13.512 & \cellcolor[HTML]{EFEFEF}\textbf{100} & \cellcolor[HTML]{EFEFEF}\textbf{0} \\ \hline
\multicolumn{1}{c|}{} & IM-GAN & 13.928 & 1.816 & 3615 & {\color[HTML]{333333} 49.64} & 41.96 & 47.79 & 58.59 & 69.05 & 68.58 & 6.298 & \textbf{100} & \cellcolor[HTML]{FFFFFF}\textbf{0} \\
\multicolumn{1}{c|}{} & SDF-StyleGAN & 15.763 & 1.839 & 3730 & 45.60 & 45.50 & 43.95 & 63.25 & 67.8 & 67.66 & 6.846 & \textbf{100} & \cellcolor[HTML]{FFFFFF}\textbf{0} \\
\multicolumn{1}{c|}{} & MeshDiffusion & {\color[HTML]{333333} \textbf{13.212}} & 1.731 & 3472 & 46.00 & 46.71 & 42.11 & {\color[HTML]{333333} \textbf{53.69}} & {\color[HTML]{333333} 57.63} & 63.02 & 5.038 & 22.12 & 62.68 \\
\multicolumn{1}{c|}{} & GenUDC & 14.083 & {\color[HTML]{333333} 1.653} & {\color[HTML]{333333} 2924} & 48.08 & {\color[HTML]{333333} 48.60} & {\color[HTML]{333333} \textbf{47.94}} & 59.18 & 58.67 & {\color[HTML]{333333} 60.84} & {\color[HTML]{333333} 4.837} & 1.47 & 46.76 \\
\multicolumn{1}{c|}{\multirow{-5}{*}{Chair}} & \cellcolor[HTML]{EFEFEF}\textbf{Ours} & \cellcolor[HTML]{EFEFEF}13.545 & \cellcolor[HTML]{EFEFEF}{\color[HTML]{000000} \textbf{1.569}} & \cellcolor[HTML]{EFEFEF}\textbf{2884} & \cellcolor[HTML]{EFEFEF}{\color[HTML]{000000} \textbf{51.70}} & \cellcolor[HTML]{EFEFEF}{\color[HTML]{000000} \textbf{49.04}} & \cellcolor[HTML]{EFEFEF}46.09 & \cellcolor[HTML]{EFEFEF}58.37 & \cellcolor[HTML]{EFEFEF}{\color[HTML]{000000} \textbf{56.70}} & \cellcolor[HTML]{EFEFEF}{\color[HTML]{000000} \textbf{60.17}} & \cellcolor[HTML]{EFEFEF}{\color[HTML]{000000} \textbf{3.935}} & \cellcolor[HTML]{EFEFEF}\textbf{100} & \cellcolor[HTML]{EFEFEF}\textbf{0} \\ \hline
\end{tabular}
}
\label{table:metric}

\end{table*}

\begin{table*}[t]
\caption{
Quantitative evaluation of shape generation at $128^{3}$ resolution on airplane category.
}
\centering

\resizebox{0.9\linewidth}{!}{
\begin{tabular}{lccc|ccc|ccc|c|c|c}
\hline
\multicolumn{1}{c}{} & \multicolumn{3}{c|}{MMD ($\downarrow$)} & \multicolumn{3}{c|}{COV ($\%, \uparrow$)} & \multicolumn{3}{c|}{1-NNA ($\%, \downarrow$)} &  &  & \multicolumn{1}{l}{} \\ \cline{2-10}
\multicolumn{1}{c}{\multirow{-2}{*}{Method}} & CD $\times 10^3$ & EMD $\times 10$ & LFD & CD & EMD & LFD & CD & EMD & LFD & \multirow{-2}{*}{JSD $\times 10^3 (\downarrow)$} & \multirow{-2}{*}{\begin{tabular}[c]{@{}c@{}}Watertight\\ ($\%, \uparrow$)\end{tabular}} & \multicolumn{1}{l}{\multirow{-2}{*}{\begin{tabular}[c]{@{}l@{}}Self-int.\\ ($\%, \downarrow$)\end{tabular}}} \\ \hline
\multicolumn{1}{l|}{LAS-Diffusion} & 4.654 & 0.56 & 3142 & 37.45 & 35.72 & {\color[HTML]{333333} 42.15} & 79.48 & 84.67 & 71.51 & 33.137 & \textbf{100} & \textbf{0} \\
\multicolumn{1}{l|}{GenUDC} & 4.000 & 0.509 & 3077 & \textbf{46.72} & 43.88 & 42.27 & \textbf{60.01} & 61.06 & 69.22 & 6.873 & 1.2 & 56.3 \\ \hline
\rowcolor[HTML]{F2F2F2} 
\multicolumn{1}{l|}{\cellcolor[HTML]{F2F2F2}\textbf{Ours}} & \textbf{3.852} & \textbf{0.500} & \textbf{2975} & 44.74 & \textbf{50.06} & {\color[HTML]{333333} \textbf{44.14}} & 62.70 & \textbf{58.52} & \textbf{68.75} & \textbf{6.510} & \textbf{100} & \textbf{0} \\
\hline
\end{tabular}
}

\label{table:128res}

\end{table*}
\begin{table*}[htb]
\caption{
Ablation study on the airplane category to evaluate the effectiveness of $D_{occ}$.
}
\centering

\resizebox{0.9\linewidth}{!}{
\begin{tabular}{lccc|ccc|ccc|c|c|l}
\hline
\multicolumn{1}{c}{} & \multicolumn{3}{c|}{MMD ($\downarrow$)} & \multicolumn{3}{c|}{COV ($\%, \uparrow$)} & \multicolumn{3}{c|}{1-NNA ($\%, \downarrow$)} &  &  &  \\ \cline{2-10}
\multicolumn{1}{c}{\multirow{-2}{*}{Method}} & CD $\times 10^3$ & EMD $\times 10$ & LFD & CD & EMD & LFD & CD & EMD & LFD & \multirow{-2}{*}{JSD $\times 10^3 (\downarrow)$} & \multirow{-2}{*}{\begin{tabular}[c]{@{}c@{}}Watertight\\ ($\%, \uparrow$)\end{tabular}} & \multirow{-2}{*}{\begin{tabular}[c]{@{}l@{}}Self-int.\\ ($\%, \downarrow$)\end{tabular}} \\ \hline
\multicolumn{1}{l|}{Ours w/o $D_{occ}$} & 4.592 & 1.491 & 3732 & 43.75 & 41.21 & {\color[HTML]{333333} 37.02} & 72.19 & 66.02 & 81.69 & 9.899 & 75.5 & \multicolumn{1}{c}{12.5} \\
\multicolumn{1}{l|}{Ours w/ $D_{occ}$} & \textbf{3.751} & \textbf{0.827} & \textbf{3007} & \textbf{49.82} & \textbf{50.95} & \textbf{44.91} & \textbf{60.01} & \textbf{55.15} & \textbf{69.26} & \textbf{7.566} & \textbf{100} & \multicolumn{1}{c}{\textbf{0}} \\ \hline
\end{tabular}
}

\label{table:ablation}

\end{table*}
\textbf{Baselines} For qualitative comparison, we compare with IM-GAN \cite{im_gan}, SDF-StyleGAN \cite{sdfstylegan}, LAS-Diffsuion \cite{lasdiffusion}, MeshDiffusion \cite{meshdiffusion}, and GenUDC \cite{GenUDC}. We do not compare with sequence-based methods such as Meshgpt, PolyGen, and PolyDiff, since they can not produce results with complex structures and have constraints on the number of faces they can handle as noted in \cite{GenUDC}.

\textbf{Metrics} We use common metrics for shape generation, which are, Minimum Matching Distance (MMD), Coverage (COV), Leave-one-out Accuracy (1-NNA), and so on. In addition, we include watertightness and self-intersection tests to demonstrate that our results retain fine details while achieving surfaces with better properties.

\textbf{Quantitative Evaluation} Quantitative results in 64 resolutions are provided in Table~\ref{table:metric}. Our method performs best on the MMD metric, exceeding the state-of-the-art GenUDC in almost every category, suggesting that QEMesh generates results closely aligned with the real data distribution. For COV, 1-NNA, and JSD metrics, QEMesh outperforms the SOTA in several categories, especially table and chair, while remaining competitive in the other categories. For watertightness and non-self-intersection, our results surpass both MeshDiffusion and GenUDC. Notice that while IM-Net and SDF-StyleGAN utilizing MC for mesh extraction are expected to preserve watertight and non-self-intersection, they may exhibit some flaws in visual quality. We also conduct experiments at a resolution of 128, and the results in Table~\ref{table:128res} show that QEMesh maintains a notable performance at higher resolution.
\newline \textbf{Qualitative Evaluation} 
The visual results of all methods in $64^3$ resolution are shown in Fig.~\ref{fig:visual}. Implicit methods often produce unnatural surface patterns or lose structural details, e.g., surfaces of cars and wings of airplanes. MeshDiffusion tends to produce over-smooth surfaces, due to its reliance on Laplacian smoothing, which leads to the loss of finer details and intricate structures such as barrels of rifles, and arms of chairs. GenUDC can effectively preserve fine details, but its outputs often exhibit prominent holes or jagged edges such as the window of the jeep. In contrast, ours can generate meshes with both detailed fidelity and watertight surfaces. 

\subsection{Ablation Study of Boolean Mask Generation}

\begin{figure}[tbp]

    \centering
    \includegraphics[width=1\linewidth]{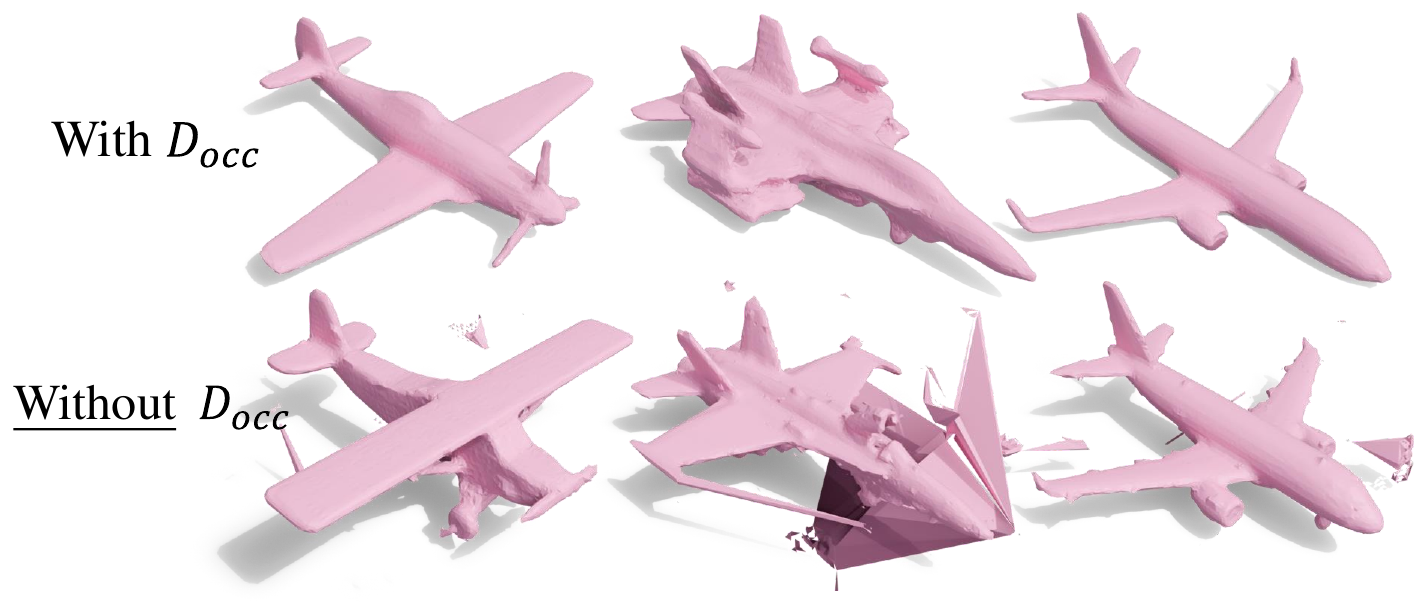}

    \caption{
    Visual comparison of results from QEMesh with $D_{occ}$ and QEMesh without $D_{occ}$.
    }
    \label{fig:ablation}
\end{figure}
To demonstrate the effectiveness and necessity of our design of the occupancy decoder, we compare QEMesh with its reduced version. In the reduced version, the occupancy decoder directly predicts the occupancy value of each point in the space. All other settings remain identical to those in QEMesh. We present samples generated by both QEMesh and QEMesh-reduced at 64 resolution in Fig.~\ref{fig:ablation}. Due to the overwhelming number of points, i.e., $\textbf{K} \times 64^3$, requiring occupancy evaluation, QEMesh-reduced often produces incorrect predictions, leading to artifacts on the surface or significant noise surrounding it. However, by predicting the occupancy value of a voxel, QEMesh can generate the boolean mask with greater accuracy and efficiency. This greatly ensures the continuity of the overall structure and the watertightness by precisely selecting the most relevant parameters. The quantitative evaluation in Table~\ref{table:ablation} demonstrates the effectiveness of our design.
\section{Conclusion}
We propose a novel mesh generation model dubbed QEMesh using the PoNQ representation. To integrate PoNQ with a generative model, we devise a multi-decoder PoNQ VAE that converts SDF to PoNQ and builds a latent space for distribution learning. The VAE consists of three parameter decoders generating PoNQ parameters and an occupancy decoder filtering the parameters contributing to the shape. Rather than predicting the occupancy of an individual parameter, the occupancy decoder predicts the occupancy of grid cells, each containing multiple parameters. This ensures that the parameters are correlated, thus maintaining structural integrity. Quantitative experiments show that QEMesh outperforms SOTA in generation quality, while ablation studies further validate the effectiveness of our decoder design.
\section*{Acknowledgment}
This work is supported by the National Natural Science Foundation of China (No. 61773270).
\bibliographystyle{IEEEbib}
\bibliography{icme2025_template_anonymized}

\vspace{12pt}

\end{document}